\pdfoutput=1

\documentclass[11pt]{article}

\usepackage{acl}

\usepackage{times}
\usepackage{latexsym}

\usepackage[T1]{fontenc}

\usepackage[utf8]{inputenc}

\usepackage{microtype}

\usepackage{graphicx}

\usepackage{amsmath,amsfonts,bm}

\newcommand{\sprice}{s_{\price}}
\newcommand{\sembed}{s_\texttt{embed}}
\newcommand{\sresp}{s_{\rtype}}
\newcommand{\sutt}{s_{\utt}}
\newcommand{\scontext}{s_{\texttt{context}}}

\newcommand{\rtype}{\texttt{type}}
\newcommand{\utt}{u}
\newcommand{\price}{\texttt{price}}

\newcommand{\aembed}{a_\texttt{embed}}
\newcommand{\aprice}{a_{\price}}
\newcommand{\aresp}{a_{\rtype}}
\newcommand{\autt}{a_{\utt}}

\newcommand{\LM}{{\textrm{LM}}}

\newcommand{\Qtarget}{Q_{\textrm{target}}}

\def\eqref#1{equation~\ref{#1}}

\def\1{\bm{1}}

\DeclareMathAlphabet{\mathsfit}{\encodingdefault}{\sfdefault}{m}{sl}
\SetMathAlphabet{\mathsfit}{bold}{\encodingdefault}{\sfdefault}{bx}{n}

\newcommand{\E}{\mathbb{E}}

\usepackage[utf8]{inputenc} 
\usepackage[T1]{fontenc}    
\usepackage{url}            
\usepackage{booktabs}       
\usepackage{amsfonts}       
\usepackage{nicefrac}       
\usepackage{microtype}      
\usepackage{xcolor}         
\usepackage{multicol}
\usepackage{wrapfig}
\usepackage{tabularx}

\definecolor{mygreen}{RGB}{10, 80, 10}
\newcommand{\User}{\textcolor{mygreen}{\textbf{Buyer:}\,}}
\newcommand{\Chatbot}{\textcolor{blue}{\textbf{CHAI:}\,}}
\newcommand{\Cocoa}{\textcolor{blue}{\textbf{Retrieval:}\,}}
\newcommand{\Supervised}{\textcolor{blue}{\textbf{LM:}\,}}

\title{CHAI: A CHatbot AI for Task-Oriented Dialogue with Offline Reinforcement Learning}

\author{Siddharth Verma  
  \qquad
  Justin Fu 
   \qquad
  Mengjiao Yang  \qquad
  Sergey Levine\\
  UC Berkeley \\
  \texttt{vsiddharth@berkeley.edu} \\ \texttt{\{justinjfu,sherryy,svlevine\}@eecs.berkeley.edu} \\
  }

\begin{document}
\maketitle
\begin{abstract}
Conventionally, generation of natural language for dialogue agents may be viewed as a statistical learning problem: determine the patterns in human-provided data and generate appropriate responses with similar statistical properties. However, dialogue can also be regarded as a goal directed process, where speakers attempt to accomplish a specific task. Reinforcement learning (RL) algorithms are designed specifically for solving such goal-directed problems, but the most direct way to apply RL -- through trial-and-error learning in human conversations, -- is costly. In this paper, we study how offline reinforcement learning can instead be used to train dialogue agents entirely using static datasets collected from human speakers. Our experiments show that recently developed offline RL methods can be combined with language models to yield realistic dialogue agents that better accomplish task goals.
\end{abstract}

\section{Introduction}
\label{sec:introduction}

Constructing fluent and intelligent dialogue agents could pave the way for intuitive interfaces and automation of human-interactive tasks. However, this requires dialogue agents that both generate fluent, natural responses and effectively pursue the goals of the given dialogue task.  A predominant approach to training dialogue agents is through supervised learning, where an agent is tasked with imitating language provided by humans. While this can provide for fluent responses, it becomes difficult to ensure that such agents systematically pursue the goals of the dialogue interaction. If we instead view dialogue as a control problem, frameworks such as reinforcement learning (RL) could allow agents to automatically optimize dialogue with respect to a task goal through a trial-and-error process and improve over human behavior.

However, implementing an RL system in practice, where an agent learns online from interacting with real humans, can be prohibitively expensive and time-consuming. This is in stark contrast to supervised learning approaches, where we can cheaply construct datasets for training imitation agents by simply logging conversations. Therefore, existing RL approaches for dialogue often rely on interacting with a learned model of a human~\citep{li2016deep,he2018decoupling}, from which experience can be generated inexpensively. However, na\"{i}ve training in this manner can result in the dialogue agent exploiting the model, which can degenerate into non-intelligible language. To mitigate this, algorithms must typically enforce strong priors to keep generated language similar to those seen in the dataset~\citep{li2016deep,jaques2019way}, or adopt dialogue management and template-based approaches which directly re-use language seen in the dataset~\citep{he2018decoupling}.

Issues such as model exploitation and distribution shift when training on static datasets are a primary concern of offline RL~\citep{levine2020offline}, and can provide a formalized approach to tackling these problems. While offline RL is motivated by scaling RL to large datasets, annotated datasets for dialogue are still small compared to the large amount of raw text datasets available today. Therefore, we propose an offline, model-free approach to dialogue generation that leverages language models. Because the size of unlabeled language datasets dwarfs that of curated datasets for dialogue, using a pre-trained language model as a central component of our method allows it to learn aspects of language fluency from unlabeled datasets, while learning higher-level strategies for goal-directed dialogue via RL on a smaller annotated datasets. This combined approach enables us to utilize the large amounts of existing language data that standard RL methods cannot.

The main contribution of this work is CHAI (CHatbot AI), an algorithm for learning task-oriented dialogue that utilizes a language model in conjunction with offline RL. We show that this leads the policy to generate goal-oriented dialogue that is both realistic and functional, and does not require training against a simulated model of human language. We evaluate our method on a negotiation task, which requires the model to both reason about strategic aspects of conversation along with generating fluent language.
We show that CHAI consistently bargains for better prices and with higher rates of successful negotiation than prior RL approaches to goal-oriented dialogue.

\section{Related Work}
Recent developments in deep learning have led to end-to-end approaches to dialogue using supervised learning, such as sequence-to-sequence models~\citep{duvsek2016sequence,eric2017copy}, hierarchical models~\citep{serban2016multiresolution}, attention~\citep{mei2017coherent,chen2019semantically}, and Transformer-based models~\citep{wu2019alternating,hosseini2020simple,peng2020soloist,adiwardana2020towards}. However, supervised learning only allows an agent to imitate behaviors, requires optimal data, and does not allow agents to exceed human performance. Supervised learning for dialogue generation also has well-known issues such as outputting commonplace responses (e.g., \textit{I do not know}) regardless of the inputs~\cite{li2015diversity}. Therefore, additional training of the dialogue agent is required for performing goal-oriented tasks.

Task-oriented dialogue has been formulated as a sequential decision making problem in a Markov Decision Process (MDP) since the 1990s~\citep{smith1994spoken,singh1999reinforcement,williams2007partially,young2013pomdp,paek2006reinforcement,henderson2008hybrid,gao2018neural,pieraccini2009we,young2013pomdp,su2015learning,chen2020airconcierge}. Dialogue is converted into abstract states and actions from which an agent is trained using reinforcement learning (RL)~\citep{eckert1997user,levin2000stochastic,chung2004developing,georgila2006user,schatzmann2007agenda,heeman2009representing,georgila2011reinforcement,su2016continuously,fatemi2016policy,asri2016sequence,zhao2019rethinking,zhang2020task,wang2020modelling}. These methods differ in how the abstract states/actions are designed and whether the simulated environment for training the policy is hand created, learned as a fixed model, or is an agent itself. For instance, \citet{eckert1997user,levin2000stochastic} learn a fixed transition model from human conversations and~\citet{georgila2011reinforcement} learn negotiation agents where each agent is the user simulator for the other agent. These methods also differ in how the decision making policy is trained, e.g., online~\citep{gavsic2011online} or  off-policy/offline~\citep{yu2016strategy,pietquin2011sample} using actor-critic~\citep{su2017sample}, policy gradient~\citep{he2018decoupling}, or fitted Q-iteration~\citep{pietquin2011sample}. Regardless of the RL method used, since policies are trained on abstract states and actions, these methods lack the ability to \textit{generate} natural language (i.e., response is created via templates depending on an abstract action). 

To overcome these limitations, recent work has trained policies directly on text, using a recurrent neural network to output language tokens, and using self-play for policy training while interacting with another learned agent~\citep{li2016deep,lewis2017deal,liu2018dialogue}.
To further improve the generated language quality, hierarchical methods decouple the strategic high-level dialogue decisions from generation~\citep{yarats2018hierarchical,he2018decoupling,saleh2019hierarchical}. These model-based approaches require accurate estimation of the environment/human (e.g., the trained self-play agent needs to mimic complex human behavior), which is beyond current capability of model-based reinforcement learning algorithms. Similar to our proposal, \citet{jaques2019way} use offline RL based on KL-control for text generation in open-domain dialogue. Our work differs in that our model is able to utilize large amounts of unsupervised data through the use of pre-trained language models, and that our work focuses on task-oriented (as opposed to open-domain) dialogue tasks. Goal-oriented tasks have clearly defined objectives that can be quantified, allowing us to provide an objective comparison between our method and prior approaches.

\section{Preliminaries}

In this section, we describe our evaluation task and cover the necessary background and notation.

\subsection{Negotiation via Task-Oriented Dialogue}
\label{sec:negotiation_task}

\begin{figure*}[h]
\small 
\textbf{Title: iPhone 5S 16 GB black silver - \$135
}

\textbf{Description:} Great condition.  No scratches.
I upgraded to iPhone 7.
Please reply with phone number if you are interested. Thank you.

\hrule 
\begin{minipage}{0.48\textwidth}
\vspace{0.2cm}

\User Hi, I would like to buy the iPhone 5S

\Chatbot Hi, great!  The phone is in great condition.  I'm asking \$135.00, it's in mint condition.

\User Does it come with a charger?

\Chatbot Yes it comes with a charger, as I upgraded. It's in mint condition.

\User How long have you been using this phone?

\Chatbot Just a month. It's in great condition and has a charger included.

\User Did you buy this phone secondhand or from the Apple store?

\end{minipage}%
\hfill
\begin{minipage}{0.48\textwidth}

\Chatbot I bought it brand new from Apple store. I only used it a month, I'm happy I did not have to use this one.

\User I can buy it for \$100

\Chatbot I'll take that if you can pick it up in a few hours.

\User Where will you be at 5pm today?

\Chatbot I'm at the Apple store so about an hour.

\User Okay, I can meet you at the Apple store.

\Chatbot Awesome!

\User \textit{accept}
\end{minipage}%

\caption{\label{fig:negotiation}An example advertisement from the CraigslistBargain task, with a conversation between CHAI (blue) and a human buyer (green).}
\vspace{-0.2in}
\end{figure*}

We evaluate our approach on the CraigslistBargain task~\citep{he2018decoupling}. CraigslistBargain consists of 6682 advertisements scraped from Craigslist, along with dialogues for each advertisement collected via Amazon Mechanical Turk where two users play the role of buyer/seller. An example advertisement from this dataset is shown in Fig.~\ref{fig:negotiation}, along with a sample conversation between a human and CHAI.

During each round of interaction, the buyer and seller can execute four possible response types. A \texttt{message} allows one player to send an utterance to the other. An \texttt{offer} allows one player to propose a price at which to conduct the transaction. Once an offer is made, the other player can either \texttt{accept} or \texttt{reject} the offer, which ends the episode. A reward is then computed based on the transaction price. Our bot receives a reward equal to the normalized price the item is sold for at the end of an episode (normalized by the list price), scaled by a constant factor of 10.
Additionally, we penalize the bot by a constant of -20 for episodes resulting in a \texttt{reject} to incentivize the agent to make deals.

We selected this task because it provides a clear objective, allowing us to illustrate our approach with quantifiable metrics. Of course, practical applications of CHAI to goal-directed dialogue may tackle other problems, including non-adversarial problms such as helping a user to answer a question or fulfill a request. However, our choice of tasks was constrained by the limited availability of public datasets for dialogue tasks that are goal-directed and have objective task goals.

\subsection{Reinforcement Learning Setup}

\begin{figure}[b]
  \begin{center}
    \includegraphics[width=1.\linewidth]{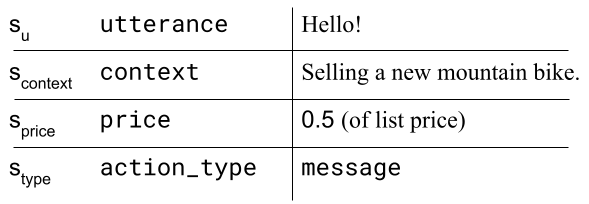}
  \end{center}
  \caption{\label{fig:statespace} An example of the state space used in the CraigslistBargain task described in Section~\ref{sec:negotiation_task}.}
\end{figure}

We formulate the task-oriented dialogue problem as an RL problem, where the agent serves the role of \textit{seller} in the CraigslistBargain problem, and the environment serves the role as the \textit{buyer}. We consider a Markov decision process defined as a tuple $(\mathcal{S}, \mathcal{A}, \mathcal{T}, r, \gamma)$. The state and action spaces, $\mathcal{S}$ and $\mathcal{A}$, consist of three main components: the action type $\rtype$ (one of the four described in Sec.~\ref{sec:negotiation_task}), an utterance $\utt$ (used only for \texttt{message} responses), and a normalized price $\price$, expressed as a fraction of the list price. Additionally, the context component $\scontext$ contains the advertisement listing description. The price component is used in two ways. First, because we do not wish to represent prices in the dialogue as discrete tokens, we instead replace prices in the utterance with a placeholder token understood to be substituted with the price component. Second, the price component is used in \texttt{offer} response type to communicate the desired transaction price. In all other cases, it is ignored. An example of the state space is shown in Fig.~\ref{fig:statespace}.

We write individual states as $s = \{\sutt, \sresp, \sprice, \scontext\}$ and actions as $a = \{\autt, \aresp, \aprice\}$, where $\sutt, \autt$ denote the \textrm{utterances}, or sequences of words generated by the environment and agent. $\{\sresp, \aresp\}$ denote the action types, $\{\sprice, \aprice\}$ denote the prices, and $\scontext$ denotes the listing context. The transition distribution $\mathcal{T}(s' | s, a)$ governs the distribution over responses generated by the buyer agent (environment), and the reward $\mathcal{R}$ defines the task objective. The goal of RL is to find a policy $\pi(a|s)$ that maximizes the expected returns:
\[ \E_{\pi, \mathcal{T}}\left[\sum_{t=0}^T \gamma^t r(s_t, a_t)\right],\]
where $r: \mathcal{S} \times \mathcal{A} \mapsto \mathbb{R}$ is the reward function, and $\gamma\in(0, 1]$ is the discount factor. In online RL, the agent is interacts with the environment to maximize this objective. In contrast, offline RL uses a dataset $\mathcal{D}$ of pre-collected interaction. This is a natural approach in many settings, such as dialogue, where online RL could require interacting with real humans for impractically long periods of time, whereas offline RL can utilize pre-recorded dialogues between humans.

\section{Offline Reinforcement Learning with Language Models}
\label{section:orl_language_models}

Offline RL potentially allows reinforcement learning methods to leverage large datasets for policy learning. However, it still requires datasets to be annotated with rewards, and for conversations to come from the task at hand. Because of this, annotated dialogue datasets, such as the CraigslistBargain dataset presented in Section~\ref{sec:negotiation_task}, are many orders of magnitude smaller than unlabeled datasets collected for unsupervised and language modeling tasks. In order to utilize these large unlabeled datasets, we propose an algorithm that combines offline RL with fine-tuned language models.

Our approach begins with training a language model, such as GPT-2~\citep{radford2019language}, and fine-tuning it on our task-specific dialogue corpus (Sec.~\ref{sec:negotiation_task}).
We use $\LM(\utt|s_{1:t})$ to denote a distribution over utterances $\utt$ produced by the language model given the dialogue history, denoted $s_{1:t}$.
We then train a critic or Q-function as described in Sec.~\ref{sec:actor_critic}, which is responsible for scoring good and bad responses and is used to select responses from a pool of candidates generated from the language model. Our approach can be viewed as using a Q-function to steer a language model (which has no concept of a task) towards producing language that accomplishes some task-specific goal.

\begin{figure*}[tb]\centering
\includegraphics[width=0.99\textwidth]{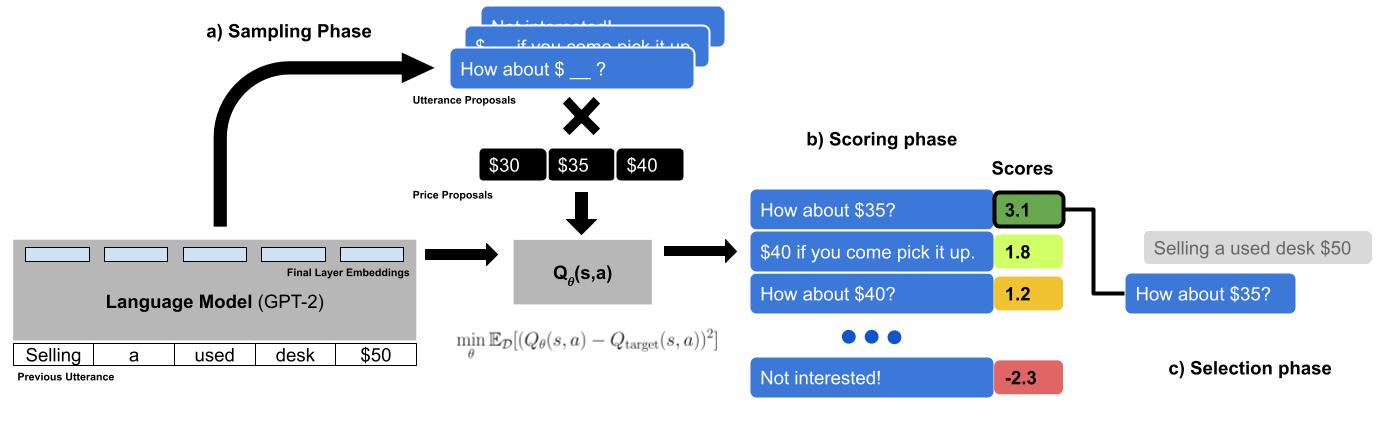}
\vspace{-3mm}
\caption{\label{fig:decoding}An overview of our dialogue generation and selection process. First (a), a finetuned language model is used to generate candidate responses conditioned on the conversation history, along with sampled prices. Then (b), each action is scored using a critic, and (c) a final response is returned by sampling the Q-value scores.}
\vspace{-5mm}
\end{figure*}

\subsection{Q-Learning with Language Models}
\label{sec:actor_critic}
In this section we describe how to train a Q-function that can score candidate responses based on their potential to maximize returns. We implement and evaluate three different training procedures, each utilizing different offline RL methods.
In the overall Q-learning framework, we sample a batch of transitions (consisting of states, actions, rewards, and successor states) from our dataset and perform updates based on minimizing a modified Bellman loss:
\begin{equation}
\label{eqn:bellman_objective}
J(\theta) = (Q_\theta(s, a) - \Qtarget(s, a))^2,
\end{equation}
where target value $\Qtarget(s, a)$ is typically computed via the Bellman operator defined as
\begin{equation}
\label{eqn:q_backup_basic}
\Qtarget(s, a) = r(s, a) + \gamma \E_{s'} \left[ \max_{a} \bar{Q}(s', a) \right].
\end{equation} 

However, using this update directly can lead to problems if we only have access to offline datasets. A widely studied issue in offline RL is the challenge of handling \emph{out-of-distribution actions}: when the maximization over the action in the target value is not constrained in any way, it is easy to obtain actions for which the Q-value predictions are erroneously high~\citep{levine2020offline}. In dialogue, this issue is greatly exacerbated, since the Q-function is only trained on responses in the dataset, and therefore is unlikely to make accurate predictions for arbitrary strings. The following modifications to Eqn.\ref{eqn:bellman_objective} and Eqn.\ref{eqn:q_backup_basic} address this issue.

\noindent \textbf{Proposal sampling (CHAI-prop)}
In the proposal sampling approach, the target value $\Qtarget(s, a)$ is computed via a modified Bellman operator that utilizes a proposal distribution based on the language model, $\mu(a_t|s_{1:t})$, to generate $N$ \emph{response proposals}, and then uses the target Q-function, $\bar{Q}$, to score those responses and selects the highest one: 
\begin{align*}
&\Qtarget^\textrm{prop}(s, a)\nonumber\\ =&\ r(s, a) + \gamma \E_{s'} \E_{\{a_i\}^N \sim \mu} \left[ \max_{i} \bar{Q}(s', a_i) \right].
\end{align*}
This sampling scheme serves a dual purpose: it both constrains the responses to be naturalistic, and it also prevents out-of-distribution inputs to the Q-function in the target value calculation. This approach resembles a number of prior offline RL methods that also employ proposal distributions~\citep{kalashnikov2018qt,kumar2019bear,fujimoto2019off,wu2019behavior}. Similarly to several prior works, we use samples from a proposal distribution for the target value, without an explicit actor~\citep{kalashnikov2018qt,ghasemipour2020emaq}. Unlike these approaches, our method leverages a pretrained and finetuned language model $\LM$, which additionally makes use of extensive unsupervised prior datasets during the pretraining stage and enables our method to handle the complex and combinatorial action space of dialogue generation. In addition, following prior work, we use a separate target network $\bar{Q}$ whose weights are updated to track those of Q using a soft update rule as done in prior methods~\citep{lillicrap16ddpg, haarnoja18sac}.

The proposal distribution $\mu(a_t|s_{1:t})$ represents a distribution over actions $a = \{\autt, \aresp, \aprice\}$. We use the language model in order to sample utterances $\autt \sim~\LM(\cdot | s_{1:t})$. To make training more computationally efficient, we pre-generate a batch of 5 utterances per transition in the dataset using the language model, and resample these during training as an approximation of directly sampling from the language model. For the prices $\aprice$, we uniformly sample a value between $70\%$ to $100\%$ of the previously offered price, which roughly matches the distribution of the seller's offers in the dataset. Finally, we infer the message type based on the utterance sampled using a simple heuristic, as the CraigslistBargain task requires us to specify a $\rtype$ for each response. During language model fine-tuning, we replaced each $\texttt{offer}$, $\texttt{accept}$, or $\texttt{reject}$ action with the utterances ``offer'', ``accept'', and ``reject'', respectively. We then check if the language model generated any of these tokens and return the corresponding action type, and label the action as a \texttt{message} otherwise. This simplifies our method and allows us to use the language model to generate the action types as well as the utterances.

\noindent \textbf{Conservative Q-learning (CHAI-CQL) }
Conservative Q-learning (CQL)~\citep{kumar2020conservative} proposes a complimentary approach to reducing the harmful effect of out-of-distribution Q-values by explicitly penalizing the Q-value of actions not seen in the dataset. We adapt CQL as an additional regularizer on the Q-value in addition to the proposal sampling scheme.
Specifically, we use the $\textrm{CQL}(\mathcal{H})$ variant, which add an additional regularizer $f^{CQL}$ to the Q-learning objective:

\begin{equation*}
\label{eqn:cql_objective}
J^\textrm{CQL}(\theta) = (Q_\theta(s, a) - \Qtarget(s, a))^2 + \alpha f^{CQL}(\theta) ,
\end{equation*}
where the regularizer is defined as:
\begin{align*}
&f^{CQL}(\theta) =\\ &\mathbb{E}_{s \sim \mathcal{D}}\left[ \log \sum_a \textrm{exp}(Q_\theta(s, a)) - \mathbb{E}_{a \sim \mathcal{D}}[Q_\theta(s, a)]  \right]
\end{align*}
Our adaptation of CQL differs from~\citet{kumar2020conservative} in that they propose an actor-critic method which trains an explicit actor. Rather, the remainder of our method is identical to the proposal sampling variant, specifically in regards to the computation of $\Qtarget$ and language model sampling. 

\noindent \textbf{Behavior-regularized Q-learning (CHAI-BRAC)}
Behavior-regularized actor-critic (BRAC)~\citep{wu2019behavior} proposes an alternative method for regularizing the Q-function such that out-of-distribution Q-values are penalized. Adapting this method to the setting of dialogue with language models, we use this approach to regularize the price proposal mechanism. Rather than uniformly sampling prices as described for proposal sampling, we train an additional \textit{price proposal network} $\pi_\phi(\aprice|s_{1:t})$ that outputs a Gaussian distribution over prices given the conversation states. Using the notation $a' \sim \pi_\phi, \mu$ to denote sampling prices from the proposal network, utterances from the language model, and action types uniformly, the proposal network is trained according to the objective
\begin{align*}
\max_{\pi_\phi} &\ \mathbb{E}_{s, a \sim \mathcal{D}} \big[ \mathbb{E}_{a' \sim \pi_\phi, \mu}[Q(s, a')] \\ 
& - D_{KL}(\pi_\phi(\cdot|s), \pi_B(\cdot|s)) \big] .
\end{align*}
The prior proposal network, $\pi_B$, is estimated as a univariate conditional Gaussian of the current offer given the previous offer, where the mean and standard deviations are linear functions of the previous offer. The target value is then computed as:
\begin{align*}
\Qtarget^\textrm{BRAC}(s, a) =\ r(s, a) + \gamma \E_{s', a' \sim \pi_\phi, \mu}\left[ \bar{Q}(s', a') \right].
\end{align*}
This adaption of the behavior regularized objective controls out-of-distribution queries on the Q-value by regularizing the price towards those seen within the dataset. This prevents the target Q-value from being queried in low-data regimes which can cause inaccuracies during training.

\subsection{Dialogue Generation} 
Once the Q-function has been trained, dialogue generation from our model is a three-phase process. The first phase is sampling: given a response from the buyer, we query the language model to sample 5 candidate utterances, and sample 5 candidate prices. The next step is scoring: we then take the cross-product of these sets, and score each potential action with the Q-function. Finally, in the selection phase, several methods are considered in order to select an action. A straightforward method is to return the action that had the highest Q-value. However, we found that this approach resulted in behavior with low diversity. Instead, we opted to follow the approach in soft Q-learning~\citep{haarnoja18sac} and sample actions from a softmax distribution over the Q-values, $p(a|s) \propto \exp\{Q(s,a)\}$, which increases diversity in the responses as sub-optimal actions are occasionally sampled. This decoding process is depicted in Fig.~\ref{fig:decoding}. 

\subsection{Architecture Details}
\label{sec:arch}

For our language model, we use an off-the-shelf implementation of GPT2-medium~\citep{radford2019language}. This language model is finetuned on a transcript of each scenario in the dataset containing the context (title, description) and spoken dialogue. The prices in the dialogue are masked out with a special price token allowing us use GPT to generate templates which we can substitute prices into. The input to the language model is a concatenation of the scenario context and the dialogue history.

The Q-function is parameterized as a feedforward network that maps states and actions into a single scalar representing the Q-value. To process the utterances into the state and action, we separately compute state and action embeddings by taking the average of the masked GPT2 attention embeddings of the entire dialogue history up to the current utterance. These embeddings are then concatenated with the prices (represented as a fraction of the list price) and message types (represented as a one-hot vector) to produce a single vector that is given to the critic as input. The critic is parametrized using a 2-layer feedforward network with hidden sizes of 256 and ReLU nonlinearities. Additional details about our model architectures, language model sampling method, and how inputs to the Q-function are structured, can be found in Appendix~\ref{app:architecture}.

\section{Experiments}
\label{sec:experiments}

\begin{table*}[t]
\centering
\small
\begin{tabular}{|l|cc|cc|cc|}
\hline
Method & \multicolumn{2}{c|}{vs Rule-based} & \multicolumn{2}{c|}{vs Stingy} & \multicolumn{2}{c|}{vs Utility} \\
& Acc\% & Revenue & Acc\% & Revenue & Acc\% & Revenue \\
\hline
CHAI-prop & $61.5$ & $0.48 \pm 0.39$ & $57.5$ & $0.39 \pm 0.35$ & $99.0$ & $0.70 \pm 0.17$ \\
CHAI-CQL & $74.0$ & $0.51 \pm 0.33$ & $77.5$ & $0.49 \pm 0.30$ & $98.0$ & $0.70 \pm 0.19$ \\
CHAI-BRAC & $62.0$ & $0.52 \pm 0.41$ & $47.0$ & $0.38 \pm 0.41$ & $99.0$ & $0.71 \pm 0.17$ \\
Language Model & $48.5$ & $0.29 \pm 0.32$ & $51.5$ & $0.27 \pm 0.28$ & $20.5$ & $0.14 \pm 0.28$ \\
\citet{he2018decoupling} (Utility) & $1.0$ & $0.01 \pm 0.10$ & $0.0$ & $0.00 \pm 0.00$ & $11.0$ & $0.07 \pm 0.22$ \\
\citet{he2018decoupling} (Fairness) & $84.0$ & $0.70 \pm 0.32$ & $80.0$ & $0.59 \pm 0.31$ & $100.0$ & $0.72 \pm 0.15$ \\
\citet{he2018decoupling} (Length) & $53.0$ & $0.46 \pm 0.43$ & $49.0$ & $0.37 \pm 0.38$ & $100.0$ & $0.72 \pm 0.16$ \\
\citet{lewis2017deal} (RL) & $83.5$ & $0.17 \pm 0.24$ & $83.0$ & $0.19 \pm 0.25$ & $64.5$ & $0.46 \pm 0.37$ \\
\citet{lewis2017deal} (SL) & $38.5$ & $0.17 \pm 0.27$ & $46.5$ & $0.21 \pm 0.27$ & $18.0$ & $0.13 \pm 0.28$ \\
\hline
\end{tabular}

\vspace{0.05in}

\begin{tabular}{|l|cc|cc||cc|}
\hline
Method & \multicolumn{2}{c|}{vs Fairness} & \multicolumn{2}{c||}{vs Length} & \multicolumn{2}{c|}{Overall (mean)}\\
& Acc\% & Revenue & Acc\% & Revenue & Acc\% & Revenue \\
\hline
CHAI-prop & $99.0$ & $0.90 \pm 0.15$ & $92.5$ & $0.79 \pm 0.27$ & $81.9$ & $\mathbf{0.65 \pm 0.34}$ \\
CHAI-CQL & $99.5$ & $0.87 \pm 0.14$ & $94.5$ & $0.79 \pm 0.24$ & $\mathbf{88.7}$ & $\mathbf{0.67 \pm 0.29}$ \\
CHAI-BRAC & $100.0$ & $0.85 \pm 0.03$ & $91.0$ & $0.76 \pm 0.25$ & $79.8$ & $\mathbf{0.65 \pm 0.34}$ \\
Language Model & $25.5$ & $0.19 \pm 0.35$ & $18.5$ & $0.14 \pm 0.32$ & $32.9$ & $0.21 \pm 0.31$ \\
\citet{he2018decoupling} (Utility) & $100.0$ & $1.00 \pm 0.00$ & $100.0$ & $1.00 \pm 0.00$ & $42.4$ & $0.42 \pm 0.49$ \\
\citet{he2018decoupling} (Fairness) & $0.0$ & $0.00 \pm 0.00$ & $100.0$ & $0.70 \pm 0.16$ & $72.8$ & $0.54 \pm 0.35$ \\
\citet{he2018decoupling} (Length) & $100.0$ & $1.00 \pm 0.00$ & $100.0$ & $0.78 \pm 0.18$ & $80.4$ & $\mathbf{0.66 \pm 0.36}$ \\
\citet{lewis2017deal} (RL) & $88.0$ & $0.26 \pm 0.34$ & $71.5$ & $0.31 \pm 0.36$ & $78.1$ & $0.28 \pm 0.33$ \\
\citet{lewis2017deal} (SL) & $60.0$ & $0.48 \pm 0.46$ & $53.0$ & $0.42 \pm 0.46$ & $43.2$ & $0.28 \pm 0.39$ \\
\hline
\end{tabular}

\caption{\label{tbl:auto_eval_results} Acceptance rates and normalized average revenue generated comparing CHAI using proposal sampling (CHAI-prop), conservative Q-learning (CHAI-CQL), and behavior regularization (CHAI-BRAC). The baselines consist of a retrieval-based agent~\citep{he2018decoupling},~\citet{lewis2017deal} and a fine-tuned language model (higher is better) against 5 different evaluation bots. The mean score across all evaluation bots is reported in the right-most column. Numbers are reported as means and standard deviations over 200 trials.
}
\end{table*}

\begin{table*}[tb]
\centering
\small
\begin{tabular}{|c|cccc|c|}
     \hline
     Metric & Fluency & Coherency & On-Topic & Human-Likeness & Total \\
     \hline
     CHAI-prop & $\bf 4.31 \pm 0.97$ & $\bf 3.91 \pm 1.17$ & $\bf 4.16 \pm 0.99$ & $\bf 3.47 \pm 1.27$& $\bf 15.84 \pm 3.86$ \\
     \citet{he2018decoupling} (Utility) & $ 3.56 \pm 1.34$ & $ 2.47 \pm 1.39$ & $ 3.09 \pm 1.40$ & $2.13 \pm 1.13$ & $11.25 \pm 4.50$\\
     Lang. Model & $4.06 \pm 1.11$ & $2.66 \pm 1.36$ & $3.63 \pm 1.18$ & $2.50 \pm 1.10$ &  $12.84 \pm 3.66$ \\
     \hline
\end{tabular}
\caption{\label{human_metric}Human evaluation scores comparing CHAI, ~\citet{he2018decoupling}, and language model (higher is better). Numbers are reported as means and standard deviations over 32 trials. CHAI scores the highest across all metrics.
}
\vspace{-3mm}
\end{table*}

Our experimental evaluation aims to compare our proposed goal-directed offline RL dialogue method to both prior dialogue management approaches and language modeling baselines. We conduct two studies: an objective evaluation against other dialogue agents to measure each method's performance in negotiation, and a subjective human study to measure the overall end-to-end performance of the system in a similar manner to prior work~\citep{he2018decoupling,jaques2019way}. Qualitative results showing actual dialogue generated from our method can be found in Appendix~\ref{app:extra_qualitative}. We consider 4 baseline approaches. The first is the current state-of-the-art approach for the CraigslistBargain task proposed by \citet{he2018decoupling} (referred to as the retrieval-based baseline). This is a hierarchical approach to dialogue generation that separately handles language generation and dialogue management. This method parses utterances into coarse ``dialogue acts,'' which represent high-level categorizations of the utterance such as greetings, offers, or counter-offers. An RL agent is then trained against a learned model of the environment to select ``dialogue acts'', and a retrieval-based generator is then used to convert ``dialogue acts'' back into text. In contrast, our method directly generates text, and does not require any manually designed categorizations of natural dialogue into dialogue acts. Since our method utilizes a modern language model, we also include a pure language modeling baseline, which consists of the same GPT-2 language model~\citet{radford2019language} finetuned on the CraigslistBargain dataset using the same method as done in CHAI. This baseline allows us to determine whether any improvement from our method is due to the language model, or to the use of offline RL. Finally, we also evaluate the end-to-end approaches described by~\citet{lewis2017deal}, which include a dialogue agent trained via supervised learning, and an RL agent optimized for the task objective.

\begin{figure*}[tb]
\small
\textbf{Title: FULLY REMODELED 4 Bdrm, 1 Bath, Duplex - \$3395}

\textbf{Description:}  4 bed / 1 bath
\textit{Type:} Duplex
\textit{City/Neighborhood:} Emeryville
\textit{Showing:} By Appointment Only
\textit{Utilities:} Paid by tenant (not included in rent amount)
\textit{Features:}
Hardwood Floor, Washer Dryer, newly renovated, stainless steel appliances
\hrule 

\begin{minipage}{.32\textwidth}
\vspace{0.2cm}

\User hi

\Chatbot Hi there. Are you interested in my rental?

\User  yes, when is the next opening to view the apartment?

\Chatbot Today.

\User how much are the utilities per month?

\Chatbot Utilities are not included in the rental amount, but I am willing to negotiate the price.

\User since utilities are not included, could you do \$3000?

\Chatbot I can't go that low. What about \$3300?

\User sounds good!

\Chatbot Perfect! Thank you!

\User \textit{accept}

\end{minipage}%
\hfill\vline\hfill
\begin{minipage}{0.32\textwidth}

\User Hi

\Cocoa hi

\User when is the next opening to view the apartment?

\Cocoa its right by apartment merrit

\User how much are the utilities per month?

\Cocoa yes they are

\User since utilities are not included, could you do \$3000?

\Cocoa deal!

\User sounds good!

\Cocoa \textit{offer \$3000}

\User \textit{accept}

\end{minipage}
\hfill\vline\hfill
\begin{minipage}{0.32\textwidth}

\User Hi

\Supervised Hi, are you interested in my property?

\User Yes, when is the next opening to view the apartment?

\Supervised Open until 6pm tomorrow.

\User Sounds good, how much are the utilities per month?

\Supervised Utilities are included in the monthly rent.

\User Okay, and can you do \$3000?

\Supervised \textit{offer \$2200}

\User \textit{accept}

\end{minipage}

\caption{\label{fig:comparison_chai_template}A comparison between similar negotiations talking to CHAI-prop (left), a retrieval-based agent \citep{he2018decoupling} (middle), and language modeling (LM, right) for the same listing. The bot takes the role of the seller, and a human is the buyer. CHAI gives both descriptive responses to questions and reasonable bargaining behavior, whereas the retrieval-based agent only shows good bargaining behavior, and the language model agent only gives descriptive responses.}
\end{figure*}

\subsection{Simulated Evaluation}

To evaluate the effectiveness of our offline RL goal-directed dialogue system, we first conducted a systematic study including the 3 variations of CHAI outlined in Sec.~\ref{section:orl_language_models}: the proposal sampling method (CHAI-prop), CQL method (CHAI-CQL), and behavior regularized method (CHAI-BRAC). In order to ensure that the results are not overfit to a single strategy, we run each method against a suite of 5 evaluation buyer agents, based on the retrieval agents presented by~\citet{he2018decoupling}. We choose these agents because they have been evaluated by humans as being human-like and have the strongest performance on the CraigslistBargain benchmark task. Specifically, we use the rule-based and RL agents proposed by~\citet{he2018decoupling} (trained using ``utility'', ``fairness'', and conversation ``length'' as rewards). To introduce additional variety in negotiation styles, we additionally modify the rule-based agent to offer 25\% of the difference between offers rather than splitting the difference between offers, which we refer to as the ``Stingy'' rule-based agent. We record the percentage of negotiations that result in an accept and the average normalized revenue generated per negotiation, which totals the average sale price (rejections have zero revenue) normalized by the listing price of the advertisement. Our results are presented in Table~\ref{tbl:auto_eval_results}.

Overall, we find that among the variations of CHAI, the conservative Q-learning variant performs the best by a small margin, but results are very comparable between all 3 variations. This suggests that the particular choice of offline regularizer is far less important than the CHAI framework of utilizing a pre-trained language model with Q-function scoring. On average, CHAI-CQL performs significantly higher on acceptance rate and similarly on revenue to the next best agent, the retrieval agent~\citep{he2018decoupling} using conversation length as reward. Computing statistical significance between these two methods, we find that $p<1.96 * 10^{-9}$ using a chi-squared test for acceptance rate, indicating that the difference in acceptance rates is statistically significant. We find that $p < 0.946$ using a t-test for revenue, indicating that the difference in revenue is not significant.
We also note that the performance of CHAI has significantly less variation across evaluations against different buyer agents than the retrieval-based agents. For example, the retrieval agent with utility reward scores near-zero on 3 evaluations but scores near-perfectly on the other two. This suggests that the CHAI framework produces dialogue agents that are more consistent and less susceptible to exploitation. We also note that implementing the retrieval method~\citep{he2018decoupling} requires hand-designing high-level dialogue actions, topic categories, and rules for parsing or labeling these components. These designs are specifically tailored to the CraigslistBargain task, whereas such hand-engineering for CHAI does not exist outside of the interface requirements to the task itself. Thus, CHAI has significantly weaker assumptions, generates dialogue end-to-end via RL, and yet is able to narrowly outperform prior methods. Among prior methods with similar assumptions to CHAI (the language modeling baseline and ~\citep{lewis2017deal}), CHAI outperforms by a wide margin on both acceptance rates and revenue.

We ran an additional ablation study on the choice of reward in Appendix~\ref{app:reward_ablation}. We find that this has a significant effect on performance, and we based our reward design on balancing between maximizing acceptance rate (through a rejection penalty) and revenue (through the utility function).

\subsection{Human User Study}
\label{sec:user_study}
To evaluate the effectiveness and naturalness of our offline RL goal-directed dialogue system, we conducted a user study with 16 individuals, who were each asked to carry out 2 negotiations with each of three agents. The users were then asked to rate the conversation on fluency, coherency, on-topicness, and human-likeness on a 5-point Likert scale. \textbf{Fluency} specifically refers to the frequency of grammatical and word-choice errors. \textbf{Coherency} measures whether the agent's responses are coherent. \textbf{On-topic} measures how well the agent was aligned with performing the task at hand. Finally, \textbf{human-likeness} measures how similar the agent's responses were to a human. Because of the cost of human evaluations, we were limited in our ability to evaluate as many baselines. Therefore, we chose methods that were the most directly comparable - the simplest variation of CHAI (CHAI-prop) optimized for utility, evaluated against the utility-optimized agent from~\citet{he2018decoupling}, and a language model baseline that shares the same finetuning procedure as CHAI. Additional details of the user study, including the questions posed to the users and statistical significance tests, are included in Appendix~\ref{app:userstudyparams}.

Results are shown in Table~\ref{human_metric}.  We ran a one-way repeated measures ANOVA test, and found that the type of agent used leads to statistically significant rating differences for all metrics (with at least $p < 0.01$). CHAI outperforms both baselines on almost all metrics, except for fluency, where both CHAI and the language model perform similarly. The fact that fluency is similar between the two models makes sense, since both methods use a GPT-2 model to generate utterances. However, the langauge modeling baseline lacks an understanding of the task goal, and therefore makes unreasonable offers or responds in illogical ways (see Appendix~\ref{app:extra_qualitative} for examples). It therefore scores lower on other metrics as compared to CHAI. This result suggests that the ability of language models to execute goal-directed dialogues is limited by a lack of awareness of the task objectives, and that offline RL potentially addresses this issue, producing dialogue that is perceived as more coherent, task-oriented, and human-like.

In Fig.~\ref{fig:comparison_chai_template} we show a comparative example between CHAI, the retrieval-based agent \citep{he2018decoupling}, and the language modeling baseline on the same scenario with human responses. CHAI and the language modeling baseline tend to produce more specific responses to the prompt due to the use of a language model, rather than generating text via the usage of templates. For example, when presented with a questioned about utilities and viewing time, both methods are able to answer the question, whereas the retrieval-based agent gives a non-sequitur answer. However, the language modeling baseline struggles in understanding prices, and offers $\$2200$ when the buyer requested $\$3000$. CHAI is able to demonstrate understanding both language and the flow of the negotiation by offering reasonable counter-offers to the user, such as responding to a low offer with ``I can't go that low'' and offering a higher counter-offer that the Buyer accepts. Additional examples from our human evaluation can be found in Appendix~\ref{app:extra_qualitative}.

\section{Discussion and Future Work}
\label{sec:discussion}

We presented a system for goal-directed dialogue based on combining offline RL with finetuned language models. CHAI learns with RL, but does so from offline datasets of human dialogue. The language model allows CHAI to benefit from large-scale unsupervised pre-training, and the offline RL component enables CHAI to select responses that are more likely to lead to a successful task outcome. Quantitatively, CHAI achieves higher acceptance rate at higher revenue than  prior dialogue management systems designed for this task.

Goal-oriented dialogue agents have many potentially useful applications such as building personal assistants, improving accessibility to technology for the disabled or the elderly, and simply saving time by automating menial tasks. Of course, as with any natural language generation technology, this kind of method can be used both beneficially and maliciously, for example by users who aim to create intentionally deceptive and realistic agents. 

While CHAI provides a proof-of-concept that offline RL can successfully learn complex human-interactive tasks such as dialogue, it also has limitations. The goal of RL is to maximize reward, which can lead to unintended responses -- for example, without additional objective terms, there is no reason for CHAI to be truthful. Similar issues affect language models more broadly, though we anticipate that it would be easier to address such issues in RL by employing better reward design. Although reward design can itself be a difficult problem, it does provide a more direct lever for influencing the agent's behavior than what is available in standard language models, which must be directed either through the choice of training data or other indirect mechanisms. In CHAI, we only investigated a single task due to architectural constraints, and the exact same architecture presented in this paper (e.g. a price prediction head) may not be directly transferable to other domains. However, using a value-based selection mechanism is more generally applicable to any goal-oriented task.

An exciting direction for future work is to extend offline RL to address a wider range of human-interactive tasks, particularly tasks with longer-range dependencies and delayed rewards, where complex task goals can lead to the emergence of dialogue that is ultimately more useful to human users.

\section{Acknowledgements}
We thank Natasha Jacques, Daniel Fried, Dilek Hakkani-tur, Yang Liu, Alexandros Papangelis, and Wei Wei for insightful discussions. We would also like to thank all of the anonymous participants in the user study. This research was supported by the Office of Naval Research.

\newpage

\bibliography{acl2021}
\bibliographystyle{acl_natbib}

\newpage
\appendix
\onecolumn
\section{Appendix}

\subsection{Architecture Details}
\label{app:architecture}

For our language model, we use an off-the-shelf implementation of GPT2-medium~\citep{radford2019language}. This language model is finetuned on a transcript of each scenario in the dataset containing the context (title, description) and spoken dialogue. The prices in the dialogue are masked out with a special price token allowing us use GPT to generate templates which we can substitute prices into. To produce language samples, we concatenate the context of the scenario along with the dialogue history and feed it to the language model. The language model then generates the next utterance. This process is repeated in order to get multiple samples.

The Q-function is a feedforward network that maps the states and actions into a single scalar representing the Q-value. To process the utterances into the state and action, we separately compute state and action embeddings by taking the average of the masked GPT2 attention embeddings of the entire dialogue history up to the current utterance. State and action embeddings are then concatenated with the prices (represented as a percentage of the list price) and message types (represented as a one-hot vector) to produce a single vector that is given to the critic as input. The critic is parametrized as a 2-layer feedforward network with hidden sizes of 256 and ReLU nonlinearities.

The equations below describe the inputs of the Q-function. ``STRCAT'' is a custom string formatting function that concatenates the context and utterances in the dialog and prefixes each utterance with the string ``Buyer:'' or ``Seller:''. ``EMBED'' calculates the masked attention embeddings from GPT2. \texttt{dialog} represents the dialog history up to the current state, $[s_{\utt, 1}, a_{\utt, 1}, \dots, a_{\utt, t-1}, s_{\utt, t}]$, and \texttt{candidate} denotes a candidate utterance $a_{\utt, t}$ generated by the language model that is being considered for scoring.

\begin{equation*}
    \begin{split}
        \sembed & = {\rm EMBED}({\rm STRCAT} (\scontext, \texttt{dialog})) \\
        \aembed & = {\rm EMBED}({\rm STRCAT} (\scontext, \texttt{dialog} +  \texttt{candidate}) \\
        s & = [ \sembed, \sprice, \sresp ] \\
        a & = [ \aembed, \aprice, \aresp ] \\
        q & = Q(s, a)
    \end{split}
\end{equation*}

\subsection{Experiment Details}
\label{app:exp_details}

\textbf{Hyperparameter Selection.} For our Q-learning algorithm, we used default hyperparameters from an SAC implementation and did not vary the parameters. We used a critic learning rate of $3*10^{-4}$, and a soft target update rate of $0.05$. For the language model architecture, we finetuned 2 GPT models architectures (GPT2-small, and GPT2-medium). In order to select which model to use, we manually rated samples from the language generated in their quality across both model types and checkpoints, and selected the best performing model. We selected the GPT2-medium architecture at training epoch 2000.

\textbf{Compute Resources.} We finetuned our language models on TPUs (TPU v3-8, 16GB memory per core with 8 cores) within a GCP instance. We trained our Q-learning models on an internal compute cluster using an Nvidia 1080 GPU (12 GB memory).

\subsection{Reward Ablation Study}
\label{app:reward_ablation}

Because we are limited in the number of evaluations possible in a human study, we use a simulated evaluation against another chatbot to run an ablation study measuring the effect of using different reward functions. Specifically, we instantiate a rule-based dialogue manager proposed by \citep{he2018decoupling} as the ``buyer'' and negotiate with it on randomly sampled scenarios from the dataset. This is done for both our method and the baselines, and the results are tabulated in Table~\ref{alg_acc_price}. 

We evaluated 5 variations of CHAI. CHAI(final) is the method used in our paper, which uses 2 components to the reward: positive reward for the price the item was sold at, and a penalty for the episode ending in a rejection. The specific reward used was 10 * the price sold (normalized by the list price) if the offer was accepted, or a penalty of -20 if the offer was rejected. CHAI(penalty) uses the same reward, except with an increased rejection penalty. CHAI(accept) is given a positive reward of +20 for episodes ending in an accept and negative reward of -20 for episodes ending in a rejection, without regard to the price. CHAI(utility) is rewarded solely for the price an item is sold at at 10 * the price sold. Finally, CHAI(fair) is rewarded for negotiating to a midpoint price between the buyer and seller's target prices.

We report the acceptance rate, average revenue, and average offers made and offers accepted. We see that CHAI(accept), CHAI(fair) and CHAI(penalty) achieve higher acceptance rates, but offer lower prices on average. In contrast, CHAI(utility) and CHAI(final) offer higher prices with lower acceptance rates, with the pure utility optimizing agent CHAI(utility) offering the highest prices with the lowest acceptance rates. Thus, we can see how changing the reward function can significantly affect the behavior of the resulting agent.

\begin{table}[h]
\centering

\begin{tabular}{|l|c|cc|c|}
\hline
Method & Accept Rate & Prices Offered & Prices Accepted & Revenue\\
\hline
CHAI(accept) & \(0.74\) & \(0.80 \pm 0.15\) & \(0.74 \pm 0.13\) & \(0.55 \pm 0.35\)\\
CHAI(fair) & \(0.76\) & \(0.80 \pm 0.15\) & \(0.75 \pm 0.13\) & \(0.57 \pm 0.34\)\\
CHAI(penalty) & \(0.90\) & \(0.77 \pm 0.14\) & \(0.77 \pm 0.12\) & \(0.68 \pm 0.26\)\\
CHAI(utility) & \(0.34\) & \(0.96 \pm 0.29\) & \(0.76 \pm 0.13\) & \(0.29 \pm 0.42\)\\
\textbf{CHAI(final)} & \(0.66\) & \(0.84 \pm 0.14\) & \(0.87 \pm 0.07\) & \(0.51 \pm 0.38\)\\
\hline
\end{tabular}
\vspace{0.5em}

\caption{\label{alg_acc_price}Mean and standard deviation of revenue, acceptance rate, and prices sold \& offered, over 50 samples when negotiating against a baseline chatbot. Revenue and prices are reported as a fraction of the list price.}
\end{table}

\subsubsection{User Study Parameters}
\label{app:userstudyparams}

\begin{figure}[h]
\includegraphics[width=0.95\textwidth]{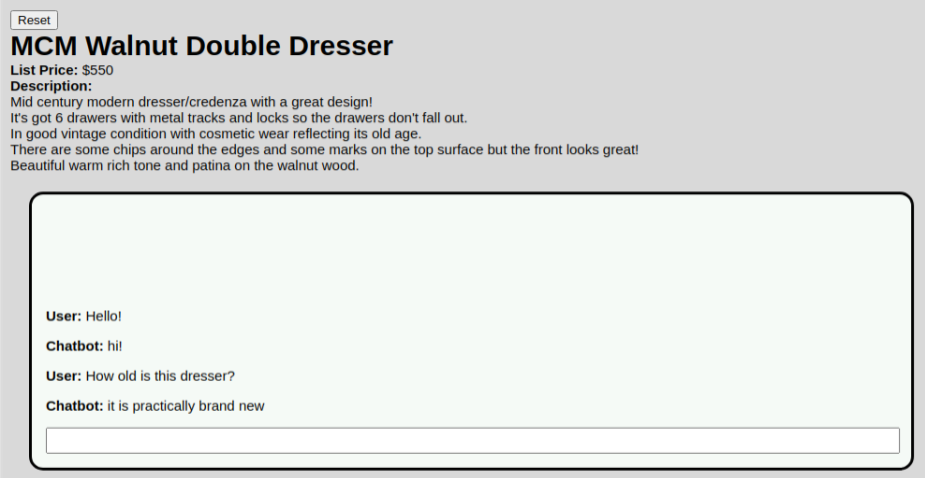}
\caption{\label{fig:user_interface}The interface presented to participants in the user study. We display the title, price, and description of the advertisement, and users are able to chat with the bot through a textbox.}
\end{figure}

\textbf{Setup.} We conducted our user study through a web-interface, where an advertisement from the test set of CraigslistBargain is displayed to the user. This is shown in Fig.~\ref{fig:user_interface}. Users were instructed to ``type any message to speak with the bot and negotiate'', and a chatbot agent replied to each message. After the user is satisfied with the negotiation, they are asked to indicate whether they want to accept or reject the current offer on the item.

After interacting with the chatbot agent, users were given a survey and asked to rate the bot on a 5-point Likert scale: strongly agree (5), agree (4), neutral (3), disagree (2), strongly disagree (1). The ratings were:

\begin{itemize}
    \item The bot was fluent (did not make grammatical or word choice errors).
    \item The flow of the conversation was coherent.
    \item The bot was on-topic.
    \item The bot demonstrated human-like behavior.
\end{itemize}
These questions correspond to the fluency, coherency, on-topicness, and human-likeness scores reported in our paper, respectively.

\textbf{Manipulated factors.} We evaluate 3 chatbot agents. The first is CHAI. The second is a retrieval-based baseline ($\textrm{RL}_{\textrm{utility}}(\textrm{act})$ as proposed in~\citep{he2018decoupling}). The this is a language modeling baseline based on finetuning GPT2-medium on the CraigslistBargain dataset.

\textbf{Dependent measures.} We measure the performance of each chatbot agent according to 4 subjective metric (fluency, coherency, on-topicness, and human-likeness), which correspond to the survey questions given to participants described above. Each metric is rated on a 5-point Likert scale. We also measure the price that was agreed upon, and whether the negotiation resulted in an acceptance or a rejection.

\textbf{Risks} The risks presented to participants were minimal, and participants were never placed in the way of physical harm. There was a small probability the chatbot could generate offensive or otherwise inconsiderate langauge, as the language generation for CHAI and the language model baseline were unconstrained. However, we did not observe this behavior prior to the study, and did not observe this behavior during the study. We minimized the risk of confidentiality breaches by anonymizing all data stored as user IDs -- participant names were not stored on our servers.

\textbf{Subject allocation.} We recruited 10 male and 6 female participants, with an average age of 24. Participation was voluntary, and subjects were not compensated for their participation. Participants were asked for consent before participating, and consent for including examples from their interactions for this paper. All examples contained in this paper are anonymized and contain no personally identifiable information.

Prior to participation in the study, each user was provided with instructions for how to use the interface, and a practice evaluation against the retrieval agent to familiarize them with the user interface and rating system. Each user interacted with all 3 dialogue agents (CHAI, retrieval, and language model) twice, with the order of interaction randomized per user. The advertisement displayed to the user is also uniformly sampled from the test set for each trial, independently from the agent being used.

\textbf{Analysis.} 
We ran a one-way repeated measures ANOVA test for each metric reported (fluency, coherency, on-topic,  human-like) to examine the effect of chatbot agent on the metric.

Results showed that the type of agent used lead to statistically significant differences in the ratings. We found that:
\begin{itemize} 
\item For fluency, $f(2, 30) = 5.8333, p < 0.0072$
\item For coherency, $f(2, 30) = 16.9518, p < 0.0001$
\item For on-topicness, $f(2, 30) = 10.1840, p < 0.0004$
\item For human-likeness, $f(2, 30) = 20.1592, p < 0.0001$
\end{itemize}

\newpage 

\subsection{Additional Qualitative Results}
\label{app:extra_qualitative}

In this section, we include additional examples collected from human dialog for CHAI-prop, the retrieval method of~\citep{he2018decoupling}, and the pure language modeling baseline.

\subsubsection{CHAI-prop}

\begin{figure}[h]

\textbf{Title: Ubiquiti Networks UAP‑AC‑LITE - \$65
}

\textbf{Description:} Price is firm.
Just a few month old.
As New. With box and all parts.  Great reliable AP. Set up once and forget about it; never need to reboot, restart like consumer AP; especially when you have high demand traffic with lots of concurrent users.  Perfect for large house or business.

\hrule 

\begin{minipage}{0.48\textwidth}
\vspace{0.2cm}

\User I noticed that your price is firm. Would you be flexible in any way?

\Chatbot That is good. I am asking \$52 for this excellent product.

\User how old is your product

\Chatbot It has been in use for only about 6 months. It has never been restarted.

\User why are you selling it

\Chatbot I do not use this product.

\User is it damaged

\Chatbot It has a small scratch on the front. It is still new. I would offer \$52.

\end{minipage}%
\hfill
\begin{minipage}{0.48\textwidth}

\User Could you do 40

\Chatbot I would be willing to let it go for \$40.

\User How about \$10

\Chatbot It is new, in excellent condition. I would not take less than \$40.

\User But you just said it was scratched

\Chatbot It is not scratched.

\User Sorry I can't do more than 35

\Chatbot Well, I can go down to \$35.

\User ok

\Chatbot Great.

\User \textit{accept}

\end{minipage}
\end{figure}

\begin{figure}[h]
\textbf{Title: Nokia Batteries - One new one used - \$10
}

\textbf{Description:} I have these two batteries left over from my old phone. Maybe you can use them? Both are in excellent working order and ready to use.
Message me if you are interested.

\hrule 

\begin{minipage}{0.48\textwidth}
\vspace{0.2cm}

\User I've been looking for batteries, what about 50 cents?

\Chatbot Hi. I still have them. They're used, but I have used them and they work perfectly.

\User How much life does the used one have?

\Chatbot \$9

\User How many times have you used the old battery?

\Chatbot 1 to 2 times

\User Did you not use your phone much?

\Chatbot Nope, I've been away from it for about 6 months.

\User Ah, I see that makes sense.

\Chatbot I can do \$6

\User I doubt many people would want to buy a used battery though, how many other offers do you have?

\end{minipage}%
\hfill
\begin{minipage}{0.48\textwidth}

\Chatbot I will let them go for \$7.

\User But you offered 6, 7 is higher?

\Chatbot No. But I'm really looking for around \$6 for them.

\User I'll pick it up right now for 4 dollars

\Chatbot No deal

\User But you were not being fair earlier when you went from six to 7 dollars

\Chatbot I can go to \$5

\User I'm holding firm on 4 dollars, especially because the batteries are used

\Chatbot I can let them go for \$5

\User No one really buys used batteries though, what other offers are you getting?

\Chatbot I'm just going to let them go for \$4

\User \textit{accept}
\end{minipage}%
\end{figure}

\begin{figure}[h]
\textbf{Title: Couch and love seat - \$80
}

\textbf{Description:} Couch and love seat from a non-smoking pets free house.
Upholstery is clean and doesn't have any stain.
Some zippers attaching the cushions to the frame are broken.
Solid wood frame.
pics coming

\hrule 

\begin{minipage}{0.48\textwidth}
\vspace{0.2cm}

\User I am interested in purchasing your couch and love seat

\Chatbot Great! These were ordered from Crate \& Barrel and are in like new condition. I'm willing to negotiate a little, but I'm asking \$72 for them.

\User How long have you owned them?

\Chatbot I've had them for a few years. They are perfect for my new apartment and they were an absolute steal when I bought them.

\User How much did you purchase them for?

\Chatbot They are brand new, I'm asking \$66.

\User How much did you originally buy the love seat and couch from from Crate and Barrel?

\end{minipage}%
\hfill
\begin{minipage}{0.48\textwidth}
\vspace{0.1in} 

\Chatbot I originally paid \$65 for both, but I figured I'd give it a shot since I love the couch.

\User So you are asking for more money then you originally bought them for?

\Chatbot Yes. I was hoping to get more for the couch and love seat. How much would you like for the couch?

\User I do, but I can only offer \$40

\Chatbot I can only go down to \$40 if you come pick up and I will throw in two chairs for you to use.

\User Great! That's a deal

\Chatbot Great!

\User \textit{accept}

\end{minipage}%
\end{figure}

\begin{figure}[h]
\textbf{Title: Fit Bike VH2 - \$200
}

\textbf{Description:} 20.75'' top tube
All in perfect working order.Very good condition - no rust - always stored indoors. Shiny. Few scratches/scrapes on pegs, etc, but very well cared for.  Gyro brake.
Thanks for interest.
If its still here it is still available.

\hrule 

\begin{minipage}{0.48\textwidth}
\vspace{0.2cm}

\User Hello, I'm interested in the bike

\Chatbot It is in great shape. It's a well-run bike. I bought it new, so I'm making it available.

\User how long have you had it?

\Chatbot I bought it new last year. It's in excellent condition.

\User can I see pictures?

\Chatbot Of course. How much are you looking to spend on the bike?

\end{minipage}%
\hfill
\begin{minipage}{0.48\textwidth}

\User I'm looking to spend under \$150

\Chatbot That's not a bad price. Would you be able to pay \$180 for it?

\User How about \$160?

\Chatbot \$160 and you have a deal.

\User Done

\Chatbot Great!

\User \textit{accept}

\end{minipage}%
\end{figure}

\clearpage
\subsubsection{Retrieval}

\begin{figure}[h]
\textbf{Title: Media Cabinet (Moving Sale) - \$20}

\textbf{Description:} Slide out tray for media storage
Condition:
Structurally sound
Good condition
Search Steve2017 for my other listings or simply
click on the "more ads by this user" link found under the map.
Please email me if you are interested.
Serious inquiries only.
Thank you

\hrule 

\begin{minipage}{0.48\textwidth}
\vspace{0.2cm}

\User Hello, how are you?

\Cocoa i am well, how are you?

\User Good. I'm interested in buying the media cabinet.

\Cocoa how can i help you today?

\User Can you provide more pictures?

\end{minipage}%
\hfill
\begin{minipage}{0.48\textwidth}

\Cocoa yes

\User How old is the item?

\Cocoa less than 1 year olf

\User I would like to buy it for \$15.

\Cocoa deal

\User \textit{accept}

\end{minipage}%
\end{figure}

\begin{figure}[h]
\textbf{Title: Adams trail-a-bike shifter 7 - \$100}

\textbf{Description:} This is an awesome bike trailer for your kid. It features 7 shiftable gears so that your kid can learn to shift but more importantly shift down or up to continue to help you pedal the bike. I know that without the shiftable gears my kid would either think the pedaling was too hard at times or going so fast they couldn't feel their pedaling helping. It took some time to teach them but definitely the way to go. 

\hrule 

\begin{minipage}{0.48\textwidth}
\vspace{0.2cm}

\User Hello, I'm wondering about the availability of this bike trailer

\Cocoa hello

\User Is this bike trailer still available?

\Cocoa i've had it for about 5 years now.

\User great, is it still available?

\Cocoa yes it is

\User When would I be able to pick it up?

\Cocoa yeah, that's fine.

\User Is there a date by which I would be able to pick it up?

\Cocoa sure, i could do that for you.

\end{minipage}%
\hfill
\begin{minipage}{0.48\textwidth}
\vspace{0.1in} 

\User Okay, is there any damage to the bike trailer

\Cocoa yes what questions do you have?

\User I'm wondering if there's any damage to the bike trailer

\Cocoa great! i'm asking \$100.00 for it!

\User Would you be able to decrease the price to \$90?

\Cocoa sure. i'd be willing to let it go for \$90.00

\User Okay great, I'll pick it up today. Does that work for you?

\Cocoa yes absolutely. its a great deal

\User \textit{accept}

\end{minipage}%
\end{figure}

\begin{figure}[h]
\textbf{Title: File Cabinet metal - \$70}

\textbf{Description:} I have 4 cabinets file metal  4-Drawer for sell. They are good condition. they are black .I sell them for 70\$ each.
Thank you for reading !

\hrule 

\begin{minipage}{0.48\textwidth}
\vspace{0.2cm}

\User Why are you selling

\Cocoa yes it is still for sale.

\User How much is it

\Cocoa it's new

\User Sorry, can you repeat that

\Cocoa where are you. how far from alameda( ca)?

\User I am close to \textit{[removed]}. It's pretty far

\Cocoa how about \$70.00?

\User I don't know the condition. Could you please provide me with some details

\end{minipage}%
\hfill
\begin{minipage}{0.48\textwidth}

\Cocoa it looks extremely heavy and in great condition! i could do \$70.00...

\User Is there any damage

\Cocoa nope. overall condition is extremely well

\User How heavy are they

\Cocoa yes, very good condition.

\User How about 25

\Cocoa deal

\User \textit{accept}

\end{minipage}%
\end{figure}

\begin{figure}[h]
\textbf{Title: Samsung Gear S2 - \$175}

\textbf{Description:} Available in Black or white
Flawless!
Gear, Samsung, S3, S2, smartwatch, smart, watch, fitbit, pedometer, fitband, fit, band, bit, Samsung Gear Fit 2, classic, iwatch

\hrule 

\begin{minipage}{0.48\textwidth}
\vspace{0.2cm}

\User Is this an S3 or S2?

\Cocoa yes. are you interested? it's in great condition

\User I'm interested.

\Cocoa do you have any questions?

\User Is it a Gear S2 or Gear S3?

\Cocoa yes its a full tablet so it does come with the charger

\User Is htis a watch or a tablet?

\end{minipage}%
\hfill
\begin{minipage}{0.48\textwidth}

\Cocoa yes.

\User Can I see pictures?

\Cocoa sure thing.

\User can I have it for \$50?

\Cocoa i am willing to do \$50.00 if you can pick it up at my place of work.

\User sure, where is that?

\Cocoa yes, that is correct.

\User \textit{reject}

\end{minipage}%
\end{figure}

\clearpage
\subsubsection{Language Model}

\begin{figure}[h]
\textbf{Title: Gorgeous 3 bedrooms available in August! Perfect way to end the summer - \$3795}

\textbf{Description:} Call Now -
Use this link for more info:
http://thegreystar.com/42v07z
Everyday Exceptional. Where You Belong.
This is a 3 Bedroom, 2 Bath, approximately 1422 Sq. Ft.
NEW APARTMENT HOMES IN PLEASANTON, CA.
A contemporary apartment community that is as much an escape as it is a destination. A place where beautifully appointed homes, resort-style amenities, rolling lawns and parks, and boutique shops come together to create a distinctly new leisure lifestyle.
Please click on the link below to check out our virtual tour!

\hrule 

\begin{minipage}{0.48\textwidth}
\vspace{0.2cm}

\User Hello, is this 3 bedroom home still available for rental?

\Supervised Hi, yes it is. The asking price is \$2430.

\User Is that a monthly cost?

\Supervised The lease is for one year, but if you have a family member or friend who are already employed and would be willing to sign a longer lease, we could offer you the option of moving in right away.

\User Would the price be lower if I signed a longer lease?

\Supervised The price is firm at \$2425, and I'm willing to negotiate

\User Would you be able to rent for \$2300?

\Supervised Yes

\User Great, what documents would I need to provide to apply to rent this place?

\Supervised This place has been carefully designed with you in mind.

\User Does this apartment have a deposit fee?

\Supervised This apartment is located in a convenient location, so that the lowest that can be reasonable for you is 10\% of your deposit.

\User What would the deposit fee bee?

\Supervised If you sign a longer lease, the deposit fee will be \$75.

\User Okay, great! How long would my lease need to be?

\Supervised I'd be willing to let it run for 5 months at no additional cost.

\end{minipage}%
\hfill
\begin{minipage}{0.48\textwidth}
\vspace{0.1in} 

\User So if I sign a 5 month lease the deposit fee would be \$75 and rent \$2300 per month?

\Supervised That's too low for me. I could come down to \$90 in monthly rent. I can also drop the rent down to \$100 for you.

\User Oh wow! So I could get this place for \$90 in monthly rent? 

\Supervised I think that is fair. How long of a lease are you looking for?

\User 1 year

\Supervised This is a very good deal. I can accept \$80 for a 3 year lease.

\User I want a 1 year lease

\Supervised Ok, you have a deal!

\User Okay, 1 year lease at \$80 per month?

\Supervised Yes

\User Great, thank y

\Supervised I would like to hear your budget.

\User \$80 per month works for me

\Supervised No, sorry.

\User I thought you had provided that as your previous offer?

\Supervised I am sorry, but I do not. So \$80 is the price.

\User Okay, confirming \$80 per month? 

\Supervised How about \$90 per month?

\User That also works.

\Supervised You have a deal!

\User Great!

\Supervised \textit{offer} \$80

\User \textit{accept}

\end{minipage}%
\end{figure}

\begin{figure}[h]
\textbf{Title: Parker Apartments - Now Accepting Below Market Rate Applications - \$979}

\textbf{Description:} ****
Parker is that thrill you feel when inspiration strikes. Perched proudly on Shattuck Avenue in
Berkeley, this collection of 155 carefully crafted, sustainably designed apartments offer immediate
access to music venues, boutique shopping, nightlife, green space, the Gourmet Ghetto, and \textit{removed}.
At home, get to know your neighbors by the crackling light of a fire pit, get focused in our
co-working space, or burn some calories at the on-site Equinox Fitness Club.

\hrule 

\begin{minipage}{0.48\textwidth}
\vspace{0.2cm}

\User Hi, when can I move in?

\Supervised Today

\User Are there grocery stores nearby?

\Supervised Yes, and the inside of our apartments is completely gated.

\end{minipage}%
\hfill
\begin{minipage}{0.48\textwidth}
\vspace{0.1in} 

\User Is there parking available?

\Supervised No, but there is a free shuttle that you can use to get there.

\User Would you be able to do \$900?

\Supervised \textit{offer} \$2800

\User \textit{reject}

\end{minipage}%
\end{figure}

\begin{figure}[h]
\textbf{Title: 20" LED Emergency Strobe Light Lamp Bar - \$30}

\textbf{Description:} 20" LED Emergency Vehicle Dash Warning Strobe Flashing Light.
these lights are a great addition and will defiantly get people out of your way as so you can respond on time!
7 Different modes of flashing / scanning mode. Please read the description to see a detail description of each mode.
Includes: 1 x Traffic strobe light bar with on/off switch and mode changer button
Fitment: Universal for all vehicle with a 12V power source
Very bright led that are visible even during daytime conditions.
2 colors: Red, Amber.
Please let me know what color you are interested.

\hrule 

\begin{minipage}{0.48\textwidth}
\vspace{0.2cm}

\User are both colors the same price

\Supervised The colors are different. I can sell them both for \$200 each.

\User I thought it was \$30

\Supervised I am sorry, I have to change my mind.

\User oops there is only one light 

\Supervised The bulbs come with the two matching colors. It would be a hassle to flash them with the other lights in your vehicle. The best thing about the bulbs is they will shine for you. It's a great deal at only \$20

\end{minipage}%
\hfill
\begin{minipage}{0.48\textwidth}
\vspace{0.1in} 

\User is there free shipping

\Supervised I think I could negotiate a little but I am not able to accept this offer.

\User can I pay \$7

\Supervised I have another offer but I can't go any lower than \$13

\User what about \$11

\Supervised \textit{offer} \$7

\User oh ok sure \$7 and free shipping?

\Supervised Yes, I accept that!

\User \textit{accept}

\end{minipage}%
\end{figure}


\end{document}